\documentclass[11pt]{article}
\usepackage{mathtools, bm, calligra, nameref, tikz}
\usepackage[linesnumbered,ruled,lined, noend]{algorithm2e}

\usepackage{enumitem}[nosep]
\usetikzlibrary{arrows.meta,
                positioning,
                shapes}
\usetikzlibrary{shapes.geometric}
\usepackage{xurl, adjustbox, realboxes, rotating, dsfont, etoolbox, nccmath, paracol, multirow, amsthm, amsmath, setspace, amssymb, fancyhdr}
\usepackage{subcaption} 
\usepackage{booktabs, makecell,afterpage, natbib, latexsym, multicol} 
\usepackage[colorinlistoftodos]{todonotes}

\usepackage{color}
\usepackage{amsmath}

\usepackage{xltabular}        
\newcolumntype{f}{>{\hsize=.2\hsize\hsize=\linewidth}X}    
\newcolumntype{s}{>{\hsize=.3\hsize\hsize=\linewidth}X}
\newcolumntype{t}{>{\hsize=.5\hsize\hsize=\linewidth}X}

\definecolor{darkblue}{rgb}{0.,0.,0.4}
\usepackage[pdftex,colorlinks=true,linkcolor=black,citecolor=black,urlcolor=black]{hyperref}
\usepackage{enumitem}
\usepackage{epsfig}

\usepackage{xcolor, graphicx, array, caption}
\usepackage[flushleft]{threeparttable}
\usepackage{newtxtext}
\usepackage{float}

\DeclareMathOperator*{\argmin}{arg\,min}

\usepackage{titlesec}
\titlespacing\section{0pt}{12pt plus 4pt minus 2pt}{0pt plus 2pt minus 2pt}
\titlespacing\subsection{0pt}{12pt plus 4pt minus 2pt}{0pt plus 2pt minus 2pt}
\titlespacing\subsubsection{0pt}{12pt plus 4pt minus 2pt}{0pt plus 2pt minus 2pt}
\titlespacing\paragraph{0pt}{12pt plus 4pt minus 2pt}{0pt plus 2pt minus 2pt}

\makeatletter
\let\@fnsymbol\@arabic
\makeatother
\makeatletter
\renewcommand*{\@seccntformat}[1]{\csname the#1\endcsname\hspace{0.2cm}}
\makeatother

\newtheorem{definition}{Definition}
\makeatletter
\g@addto@macro{\definition}{\upshape}

\patchcmd{\thebibliography}
  {\settowidth}
  {\setlength{\itemsep}{0pt plus 0.1pt}\settowidth}
  {}{}
\apptocmd{\thebibliography}
  {\normalsize  }
  {}{}
\allowdisplaybreaks[1]
\usepackage{xurl}
\usepackage{hyperref}

\usepackage{esvect}
\usepackage{scalerel}

\newtheorem{examplex}{Example}
\makeatletter
\g@addto@macro{\examplex}{\upshape}

\usepackage{todonotes, etoolbox}

\begin{document}

\title{Enhancing Supply Chain Resilience: A Machine Learning Approach for Predicting Product Availability Dates Under Disruption}

\author{Mustafa Can Camur   \thanks{General Electric  Research Center, Niskayuna, NY 12309.} \and Sandipp Krishnan Ravi $^1$ \and Shadi Saleh        \thanks{General Electric Gas Power, Atlanta, GA 30339.}}

\maketitle
\vspace{-2pc}
\begin{abstract}
The COVID-19 pandemic and ongoing political and regional conflicts have  a highly detrimental impact on the global supply chain, causing significant delays in logistics operations and international shipments. One of the most pressing concerns is the uncertainty surrounding the availability dates of products, which is critical information for companies to generate effective logistics and shipment plans. Therefore, accurately predicting availability dates plays a pivotal role in executing successful logistics operations, ultimately minimizing total transportation and inventory costs. We investigate the prediction of product availability dates for General Electric (GE) Gas Power's inbound shipments for gas and steam turbine service and manufacturing operations, utilizing both numerical and categorical features. We evaluate several regression models, including Simple Regression, Lasso Regression, Ridge Regression, Elastic Net, Random Forest (RF), Gradient Boosting Machine (GBM), and Neural Network models. Based on real-world data, our experiments demonstrate that the tree-based algorithms (i.e., RF and GBM)  provide the best generalization error and outperforms all other regression models tested. We anticipate that our prediction models will assist companies in managing supply chain disruptions and reducing supply chain risks on a broader scale.\end{abstract}  
\noindent \textbf{Keywords:} Inbound Logistics; Supply Chain Disruption;  Machine Learning;  Regression Models; Tree-based Ensemble Models

\section{Introduction} \label{Introduction}

\begin{figure}[!htbp]
\centering
     \caption{Gas turbine electrical power generation value statistics $\%$ market share (2020-2029)  \citep{Worldwid3}}
     \includegraphics[page=1,width=1 \linewidth,keepaspectratio=true]{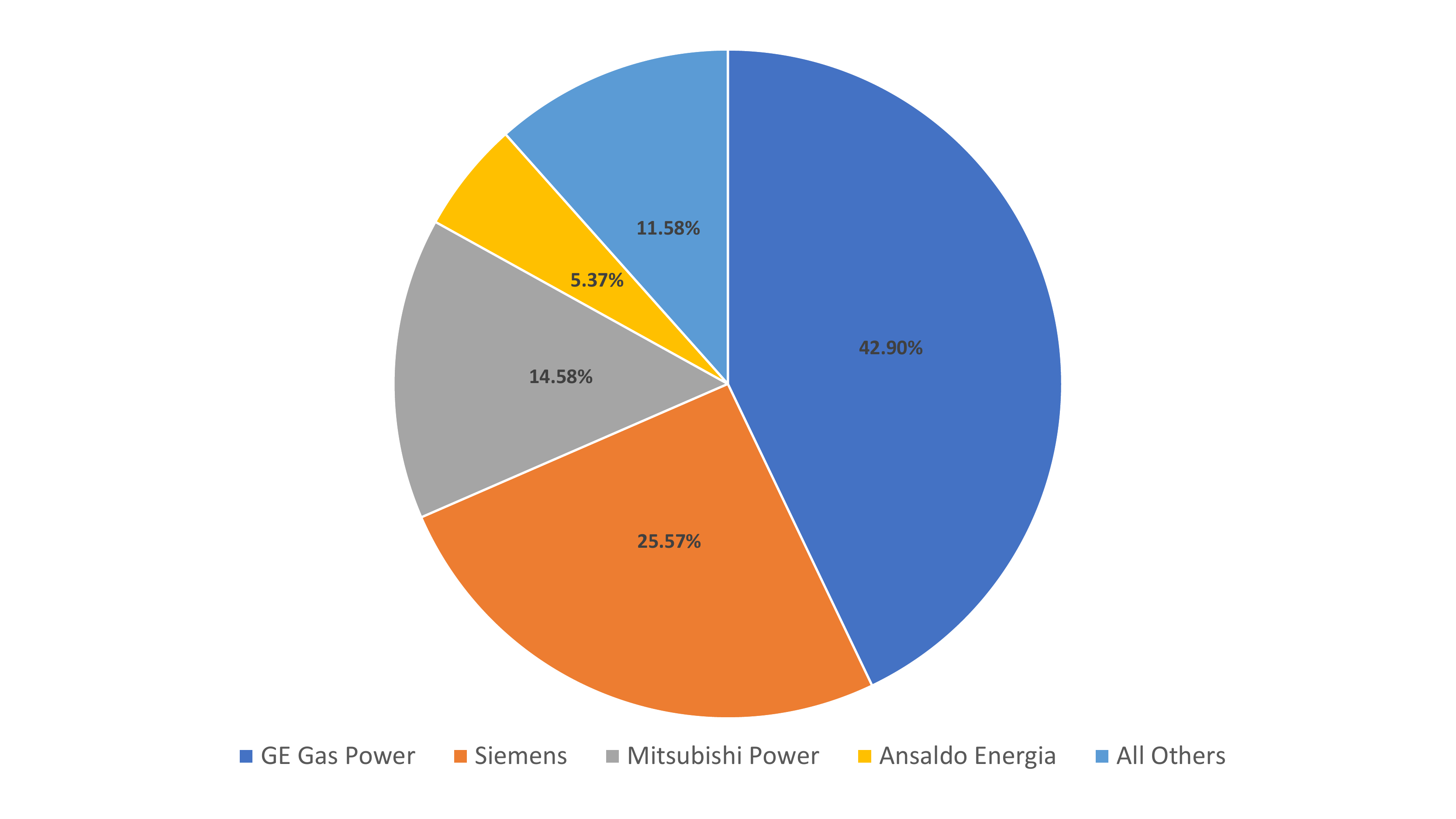}   \label{Fig2}
\end{figure}

General Electric (GE) Gas Power is a preeminent global leader in gas power technology, services, and solutions, commanding nearly 50$\%$ of the market share (see Fig.~\ref{Fig2}). Through  innovation and continuous collaboration with its clients, the company designs cleaner and more accessible energy solutions that people rely on, fostering growth and prosperity worldwide \citep{mukherjee2021ge}. With the largest installed base of gas turbines on the planet, GE Gas Power offers advanced technology and unparalleled expertise to construct, operate, and maintain gas power plants. As the world transitions towards a lower carbon future, the business is working closely with its power generation clients to help them succeed in the present and plan for the future. It is strongly believed that pre- and post-combustion decarbonization in gas turbines is crucial to decarbonizing the power sector globally.

Despite the fact that the plants responsible for manufacturing gas and steam turbines are situated within the United States (U.S.), a significant proportion of the products required for the manufacturing process are procured from suppliers located in Asia, including but not limited to China, Singapore, and South Korea. These inbound logistics shipments between GE's suppliers, plants, and warehouses are planned based on product schedule requirements. However, GE is only notified of product availability dates for pick-up from suppliers 24 to 48 hours in advance, with no means to anticipate or predict availability dates further in advance (e.g., 6 or 12 months ahead). As a result, GE is vulnerable to the spot market's high transportation costs and limited alternate mode flexibility. Additionally, uncertainties surrounding availability dates and last-minute information flows have a detrimental effect on inventory and financial planning operations.

\begin{figure}[!htbp]
\centering
     \caption{Visualization of the logistics operations via transportation mode optimization and container shipment consolidation}
     \includegraphics[page=1,width=1 \linewidth,keepaspectratio=true]{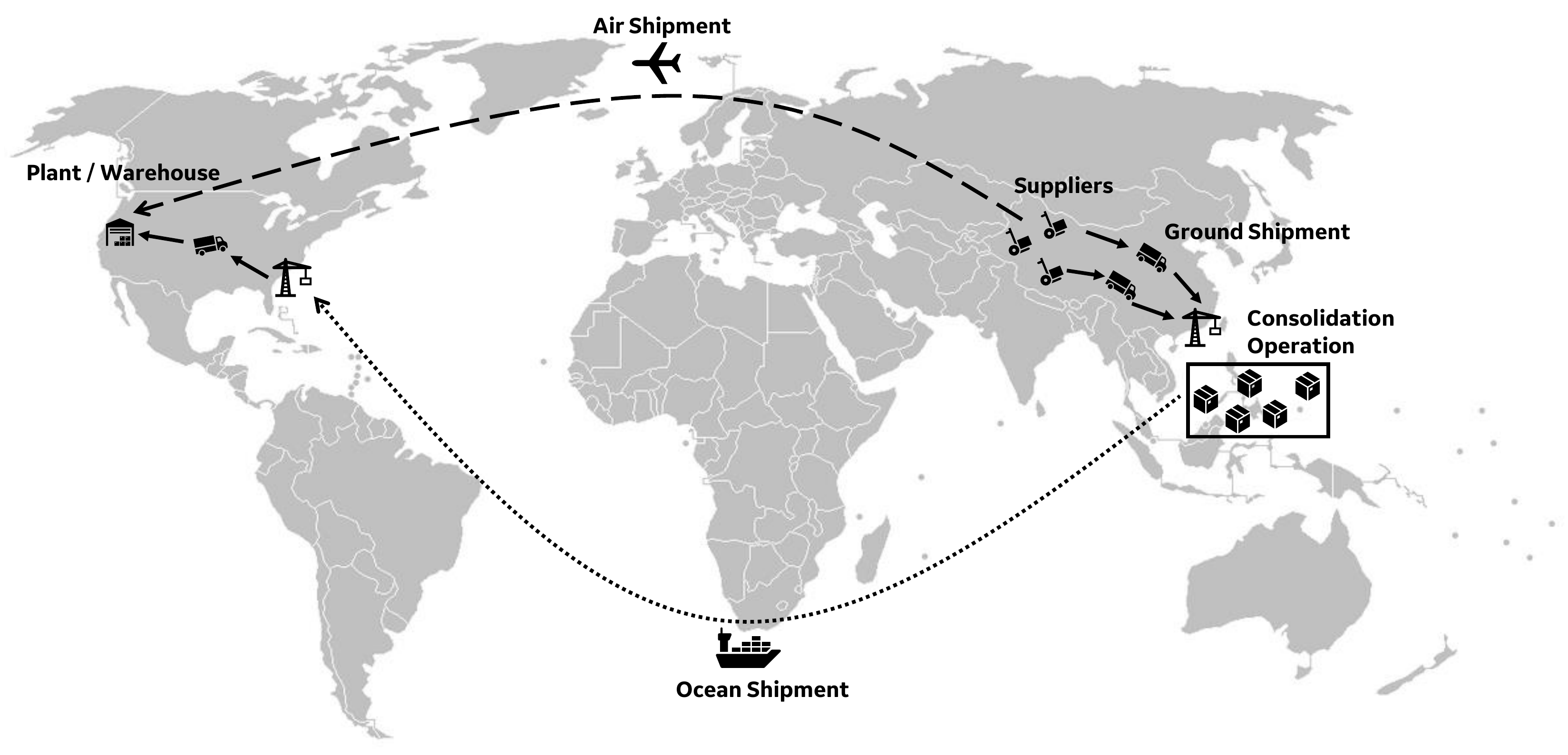}   \label{Fig1}
\end{figure}

At the GE Research Center, researchers have collaborated with the GE Gas Power logistics team to develop a mixed-integer linear programming model and a novel heuristic approach for optimizing transportation mode selection and container shipment consolidation, thereby facilitating efficient and sustainable logistics operations (see \cite{camur2022optimization}). The  model is designed to generate an annual logistics plan containing routing decisions for every product transported via different modes of transportation, such as ocean, air, and ground. The model makes operational decisions about when (i.e., which week) and where (i.e., which port location) to book a full container load (FCL) and which parts to include in each container. The objective is to minimize the total logistics costs incurred while ensuring that parts arrive at the warehouses at the desired time. We depict the problem definition in Fig.~\ref{Fig1} below and refer the reader to refer to \cite{camur2021large} and \cite{camur2022stochastic} for additional information on  optimization studies.

It is crucial to note that optimization models are highly sensitive to the model parameters. This means that the decisions made by the model may vary significantly based on the input provided. One of the most critical parameters used in our optimization model is the availability dates of products from suppliers. In order to make feasible routing decisions for a specific product, decision makers need to know when it can be picked up from the supplier. The availability dates provided by suppliers have a significant impact on a) routing decisions, b) transportation lead times, and c) logistics costs. High lead times force companies to keep buffer inventories at their warehouses to ensure that manufacturing processes are not delayed or halted, resulting in increased inventory and logistics costs \citep{chung2018investigating}.

Having a system that can predict product availability dates a couple of months in advance would have significant benefits for the GE Gas Power logistics and inventory teams. With this information, the logistics team can level load shipments, obtain better shipping rates, and consolidate assets, which can result in cost savings. The inventory team can also forecast inventory balances with greater accuracy, which can aid in operational and financial planning. Additionally, having accurate supplier and commodity lead time forecasts can help adjust material resource planning and improve overall supply chain management. Therefore, the development of a prediction system for product availability dates would be valuable for the company's operations and financial planning. 

We also note that GE announced a new organizational structure. GE Power, GE Renewable Energy, GE Digital, and GE Energy Financial Services will now be known as \textit{GE Vernova} and will be spun off by 2024 (see \href{https://www.gevernova.com/}{https://www.gevernova.com/}). As a result, we expect our work to have a broader impact on GE's portfolio of energy businesses in the future.
 
In this work, we test different regression models to accurately predict the availability date for each product order, with the intent of utilizing these predictions in an optimization model. The necessity for these predictions arises from the fact that availability times promised by suppliers, and required by the business, are generally not met due to the stochasticity faced in supply chain systems \citep{yazdekhasti2021multi}. In fact, a high percentage of orders are delayed, especially after the global supply chain disruptions experienced during the COVID-19 pandemic and the Russia-Ukraine war \citep{choi2022reforming}. As an example, all logistics operations were halted at the Port of Shanghai after its lockdown in 2022 \citep{HowShang29}.

Our work is outlined as follows. We first provide a literature review in Section \ref{Literature Review}. We then discuss our dataset and the features used during prediction in detail in Section \ref{Data Set Discussion}. Next, we introduce the machine learning (ML) algorithms that were tested to estimate availability dates in Section \ref{Methodology}. Following that, we conduct our computational experiments and compare each regression model based on different statistical key performance indicators (KPIs) (see Section \ref{Computational Experiments and statistical analysis}). Lastly, we summarize our findings and share potential future research directions in Section \ref{Conclusion}.

\section{Literature Review} \label{Literature Review}

In this section, we review the papers that are most relevant to our work. Given the global supply chain disruption experienced after 2020, there have been a plethora of studies examining supply chain resilience. The areas investigated in the context of supply chain resilience range from healthcare \citep{sazvar2021capacity} to food \citep{van2020agent}, from manufacturing \citep{rajesh2021optimal} to fashion \citep{choi2022reducing}. While \cite{remko2020research} discuss the potential research fields to increase supply chain resilience after the global pandemic, \cite{spieske2021improving} provide a holistic review to achieve a similar goal through Industry 4.0. There are a fair number of review and survey papers that might interest the reader, some of which include \cite{ribeiro2018supply}, \cite{hosseini2019review}, and \cite{katsaliaki2021supply}.

Supply chain and logistics are well-studied in both operations research and optimization communities \citep{
 suryawanshi2022optimization}. Since our focus is on machine learning (ML), we will only cover a small portion of these studies.  \cite{ben2011robust} implement a robust optimization method to satisfy uncertain demand faced in humanitarian supply chains. The authors run several simulation scenarios to conduct stress testing.\cite{aqlan2016supply} propose a hybrid method combining simulation and optimization to offer risk mitigation strategies in a high-end server manufacturing environment. The framework designed helps the decision-maker to allocate  sources at hand to meet the customer demand under a stochastic environment.  \cite{sherwin2020identifying} utilize logic and probabilistic technique (i.e., fault tree)   for the bill of materials sourced from external suppliers. The objective is to identify which supply chain structure might bring up the biggest supply chain risk and the highest cost.  They examine both perfect and imperfect mitigation models and find out that the imperfect model is easier to adopt.

\cite{liu2022optimization} study the supply chain viability problem under a limited intervention budget. The authors propose two non-linear programming models which are then solved through a genetic algorithm.   \cite{camur2021optimizing} develop an optimization model for a mass rescue event in Arctic Alaska with dynamic logistics decisions, considering scarce resources such as food and shelter. They address important policy questions that could assist in future rescue missions in the region.  \cite{sawik2022stochastic} analyze the ripple effect in supply chain operations by proposing  a multi-portfolio approach and scenario-based stochastic mixed-integer linear programming model. The author recommends pre-positioning buffer inventories and obtaining backup suppliers as mitigation strategies under stochastic disruptions.  

Interest in  machine learning (ML) applications in supply chain and logistics have been growing exponentially in literature (see \cite{ni2020systematic} for an extensive survey). \cite{ganesh2022future}  present a survey study where they discuss in detail the papers focusing on the role of ML and artificial intelligence (AI) in supply chain risk management.  They also identify gaps and unexplored areas in the literature as future research directions. \cite{carbonneau2008application} investigate ML methods such as support vector machine (SVM) and recurrent neural network to estimate the total demand for products after a bullwhip effect. The authors mention that the advanced methods do not provide a significant improvement in the generalization error compared to the simple regression models. In a similar setting, \cite{zhu2021demand} work on a demand forecasting problem in the pharmaceutical supply chain. They design a new demand forecasting structure that acquires time series data through different products and uses pattern recognition algorithms. The computational experiments show that downstream inventory information plays a significant role in demand forecasting. \cite{nguyen2021forecasting} analyze time series data to identify anomalies in sales in a fashion industry application. The authors utilize Long Short Term Memory (LSTM) networks and an LSTM Auto-encoder network strengthened by a one-class SVM algorithm to estimate time series data and identify anomalies, respectively. Further, \cite{alves2022multi} use a deep reinforcement learning (i.e., Proximal Policy Optimization) to solve a production planning and distribution problem in a multi-echelon system. They design the problem using Markov-decision process and non-linear optimization to capture uncertainty in lead times.

\cite{priore2019applying}  implement an inductive learning algorithm to cope with the fluctuating supply chain conditions (e.g., price changes, varying demand signals). The algorithm aims to select a new product replenishment policy under changing conditions to minimize the total operational costs. \cite{islam2020prediction} use tree-based learning algorithms including Distributed Random Forest (DRF) and Gradient Boosting Machine (GBM) to predict the back order of products. Since tree-based ensemble methods suffer from space complexity, the authors use a ranged-based clustering method for dimensionality reduction. The vital features used during prediction include inventory levels, lead times, sales numbers, and forecasted sales. In another study, \cite{punia2020deep} propose a hybrid approach combining both LSTM networks and a random forest algorithm for a multi-channel retailer to forecast sales demand. While they utilize features like price, discount, and brand, the approach proposed provides the best generalization error compared to the other approaches (e.g., feed-forward neural networks, multiple linear regression). \cite{kosasih2022machine} examine a supply chain visibility problem under disruption and utilize graph neural networks (GNN) to estimate the hidden links between the suppliers. They test their algorithm on a  real-world  automotive supply chain network consisting of over 18,000 suppliers and the results indicate that GNN outperforms the well-known link prediction heuristics (e.g., Preferential Attachment).

The lead times are expected to have a high correlation with the availability dates of products, and there have been a significant number of studies conducted in this area. While \cite{bender2021prototyping} design an AutoML system integrated into an ERP system to obtain high-quality predictions for customizable orders, \cite{dosdougru2021novel} propose a novel hybrid AI-based decision support system to forecast supplier lead times. To our knowledge, the study most similar to ours was conducted by \cite{banerjee2015hybrid}. The authors propose a hybrid statistical approach containing a step-wise regression-generalized multivariate gamma distribution model to forecast delivery times for aircraft engine parts. However, they did not utilize any advanced ML techniques and did not incorporate categorical data during the learning phase. Thus, we believe that our study fills an important gap in the supply chain risk literature and provides useful future directions for both academics and practitioners. Our prediction framework can be adapted to any supply chain network, regardless of commodity type, region, or sector. With the growing interest in using ML and AI techniques to revolutionize supply chains \citep{mohamed2020machine}, we humbly believe that our work may shed some light on the future of the global supply chain.

\section{Data Set Discussion} \label{Data Set Discussion}

Historical data regarding availability dates is stored in Oracle Transportation Management (OTM), a service provided by Oracle for companies to manage their supply chain operations. We identified ten features to predict the availability date of each product order, which are discussed in detail in Section \ref{Features}. While availability dates are stored in date format (e.g., mm/dd/yyyy), we convert each of these dates to a numerical value representing the number of days between the product order creation date (POCD) and the availability date. We aim to predict how many days after POCD each product order will be ready.

 \subsection{Features} \label{Features}
In this section, we elaborate on the features that are used to predict the availability date of each product. It is important to note that we are not able to share certain data and details due to confidentiality reasons. The list of the features including the data type is shared below.

\begin{itemize}
    \item Part Number (Categorical): A categorical feature that groups products based on their specifications, and provides unique information about each product and its content.
    \item  Supplier Code (Categorical): A unique identification number assigned to each supplier the business works with. While the supplier name could be used as an alternative, it is not always stored in a consistent manner in the database and may have variations in its string representation. For example, the name of a company located in China may be stored in both Latin and Chinese alphabets. Moreover, some company names may change over the years, but the supplier code remains unchanged.
    \item Supplier Location (Categorical): The country and city name of each supplier are used to measure the impact of their origin location on the availability dates. It is expected that suppliers located within the same area would exhibit similar behavior. To avoid confusion arising from similar city names in different countries (e.g., Waterloo, Belgium and Waterloo, Canada), we combine the two strings to provide better insight.
    \item Product Cost (Numerical): The total cost of each product is stored in U.S. dollars. We analyze the impact of a product's price on the supplier's ability to prepare the product. It is expected that higher-cost shipments will be ready more quickly.
    \item Product Amount (Numerical): The product orders contain information on the quantity of each product. It is expected that larger quantities may face delays during the production phase on the supplier side, thereby negatively impacting the availability dates.
    \item Product Details (Categorical): Products are classified into different subgroups based on their features. For instance, a product might belong to the electronic systems group in the first tier and the electrical group in the second tier. The product details variable is used to determine whether the product category has an impact on the availability date.
    \item Contract Delivery Time (Numerical): Upon signing a contract with a supplier, the supplier is expected to provide an approximate lead time in days. This feature is included to measure the impact of the contractual agreement on the availability dates. However, there are two issues worth mentioning. Firstly, many suppliers may choose not to report a delivery time in order to avoid penalties. Secondly, the lead times reported by suppliers may not always be accurate.
    \item Latest Need by Date (Numerical): The business shares a date for every product order when the product is needed at the latest. We calculate the number of days between the product order creation date (POCD) and the latest need by date, which is then used as a numerical variable in the predictions.
    \item Latest Promised Date (Numerical): Each supplier provides the information of when a product would be ready to be picked the latest. We count the number of days between the POCD and the Latest Promised Date, which is then used as a numerical predictor. 
    \item Approval Date (Numerical): Each product order requires an approval date, and it is expected that an earlier approval date may provide insight into the urgency or importance of the order, which in turn may lead to earlier readiness. To capture this information, we calculate the difference between the product order creation date and the approval date and use it as a numerical variable in our predictions.
\end{itemize}



\section{Methodology} \label{Methodology}

In machine learning, there are response and predictor variables that correspond to the product availability date and the features introduced in the previous section, respectively. We tested a number of regression models, including Linear Regression, Lasso Regression, Ridge Regression, Elastic Net Regression, Random Forest, Gradient Boosting Machine, and Neural Networks, to predict the product availability dates. All of these techniques are commonly used in both industry and academia. Each ML method tested is summarized below to provide a better understanding for the reader.
\subsection{Linear Regression} \label{Linear Regression}

In linear regression, it is assumed that the relationship between the response (i.e., dependent variable) and predictors (i.e., independent variables) is linear. This means that the response variable can be expressed as a linear combination of the predictor variables, plus some noise or error term. Let $n$ be the number of data instances and let $d$ be the dimensionality of the feature vector for each instance. For the $i^{th}$ data instance, we can express the relationship between the response and predictor variables as follows:

\begin{equation}
    y_i =  b + w_1 x_i^1 + w_2 x_i^2 + \cdots +  w_d x_i^d \quad \forall i= 1\cdots n
\end{equation}

where $b$ is the bias/noise value that represents the unobserved data. The purpose is to solve the linear equation above (i.e., $Y = Xw + b$ in a matrix form) to get the values for the weight vector $\vec{w}$ and bias $b$ that defines what is called a fitted line. This is typically done using gradient descent or closed-form solutions such as the normal equation. The resulting model can then be used to predict the response variable for new data instances. 

Let $\vec{x_i} = [1,  x_i^1, \cdots,  x_i^d]$ and $\vec{w}  = [w_0,   w_1, \cdots,  w_d]$ be ill-defined where $w_0=b$.  The problem solved  is defined as the minimization of the least squared errors shown below. In a more generic way, these problems are called ``loss function" represented by $\ell$. 

\begin{equation}
   \ell=  \min \sum_{i =1}^{n} (w^Tx_i - y_i)^2
\end{equation}

\subsection{Lasso Regression} \label{Lasso Regression}
In general, linear regression models are susceptible to overfitting and can be highly sensitive to outliers in the data. Lasso regression, also known as Least Absolute Shrinkage and Selection Operator, addresses overfitting by utilizing a "shrinkage" strategy \citep{ranstam2018lasso}. It is a form of $L_1$ regularization, which imposes a penalty as the sum of the absolute values of the weight values that define the fitted line.  The objective of the problem is to minimize the sum of squared errors between the predicted and actual response variables, subject to the constraint that the sum of the absolute values of the weights does not exceed a specified threshold determined by the parameter $\lambda$ as shown below. 

\begin{equation}
   \ell = \min \sum_{i =1}^{n} (w^Tx_i - y_i)^2 + \lambda \sum_{j=0}^{d} |w_j|
\end{equation}

Typically, the value of $\lambda$ is chosen through cross-validation. Higher values of $\lambda$ lead to smaller weight values and lower variance, which can result in some weight values being set to zero, making Lasso regression useful for feature selection by eliminating "redundant" features. 

\subsection{Ridge Regression} \label{Ridge Regression}

Ridge regression is a commonly used method for addressing the issue of high correlation among independent variables in linear regression models. When independent variables are highly correlated, the coefficient estimates in the model can become unstable and highly sensitive to small changes in the data. Ridge regression helps to mitigate this problem by introducing a regularization term that adds a penalty to the sum of squares of the model coefficients, thereby shrinking the estimated coefficients towards zero \citep{marquardt1975ridge}. 

Ridge regression is a type of $L_2$ regularization, which involves adding a penalty term to the objective function that is proportional to the sum of squares of the model coefficients. The objective function for ridge regression can be expressed as follows:

\begin{equation}
\ell = \min \sum_{i=1}^{n} (w^Tx_i - y_i)^2 + \lambda \sum_{j=0}^{d} w_j^2
\end{equation}

Here, $\lambda$ is a penalty term that helps control the impact of the regularization term on the overall objective.

\subsection{Elastic Net} \label{Elastic Net}

Although Lasso and Ridge regressions employ different regularization strategies, they both aim to decrease variance to improve generalization. Elastic net combines both models by taking convex combinations of the second components of the objective function (see \cite{camur2022star} for more details on convexity). The resulting model is presented as follows \citep{zou2005regularization}:

\begin{equation}
    \ell = \min \sum_{i =1}^{n} (w^Tx_i - y_i)^2 + \lambda ( \alpha \sum_{j=0}^{d} |w_j| + (1-\alpha) \sum_{j=0}^{d} w_j^2)
\end{equation}

The coefficient $\alpha$, which is user-defined and ranges between 0 and 1, balances the $L_1$ and $L_2$ regularizers. Overall, Elastic Net is a flexible and powerful method that combines the strengths of both Lasso and Ridge regressions.

\subsection{Random Forest} \label{Random Forest}

Random forest is an ensemble learning method that generates a set of regression trees, called a forest, by randomly selecting subsets of input data and features \citep{biau2016random}. Each tree in the forest is constructed independently of the others, and the randomization operation, called bootstrapping or bagging, helps reduce variance. During the construction of each tree, the algorithm selects a random subset of features at each node and chooses the best feature to split the data based on a predefined criterion (such as the reduction in variance). The final prediction is made by averaging the results returned by each tree in the forest.

Unlike the regression models discussed earlier, tree-based models do not require normalization of the data and have several hyperparameters to tune (e.g., the number of trees, the maximum depth of each tree, and the number of features to consider at each split). As a result, these methods have higher computation power and space requirements. Moreover, random forest provides a natural solution to both bias and overfitting issues by combining multiple decision trees. It also helps detect feature importance by computing the average decrease in impurity (such as the Gini index or entropy) over all the trees for each feature.

\subsection{Gradient Boosting Machine} \label{Gradient Boosting Machine}
Gradient Boosting Machine (GBM) is an extension of decision trees that aims to combine weak learners to obtain a strong learner in an iterative manner \citep{natekin2013gradient}. Unlike random forest, which reduces error by reducing variance, GBM increases the variance over time using weak learning. The algorithm works by minimizing a loss function $\ell$ that measures the difference between the predicted values and the true values of the target variable. Let $\mathcal{H}$ represent the set of all weak learners, and we solve the following optimization problem to pick the next learner at a certain iteration:

\begin{equation}
h = \argmin_{h \in \mathcal{H}} \ell (H + \alpha h)
\end{equation}

where $H(x) = \sum_{j=1}^{t} \alpha h_j(x)$ is the current model, and $\alpha$ is a learning rate that controls the contribution of the new learner. It is worth noting that GBM has a higher number of hyperparameters than random forest due to the gradient operations, which require additional parameters such as the step size and learning rate.

\subsection{Neural Network} \label{Neural Network}

Neural networks (NNs) are complex computational models that consist of multiple layers of interconnected nodes, each of which performs a nonlinear transformation on its inputs \citep{anderson1995introduction}. The connections between nodes, called edges, have associated weights that are adjusted during training to minimize a chosen loss function. This optimization is performed using backpropagation, a technique that calculates the gradient of the loss function with respect to the network's parameters. NNs have achieved state-of-the-art performance in various fields, such as image recognition and natural language processing \citep{traore2018deep}.

In this study, we focus on the traditional feed-forward NN architecture. Our objective is to minimize the mean squared error loss function, which is similar to the one used in linear regression (see Section \ref{Linear Regression}). Since we have categorical variables in our dataset, we use one-hot encoding to convert them to numerical values during the training process. One-hot encoding creates a binary vector for each categorical variable, where each element in the vector represents a unique category.

\section{Computational Experiments}\label{Computational Experiments and statistical analysis}

We conduct all experiments using the H2O library in the Python API on a desktop with an Intel(R) Xeon(R) Silver 4214R CPU at 2.40GHz and 128 GB of RAM. To prevent overfitting, we utilize regularization techniques and 5-fold cross-validation. The $\lambda$ parameter for both Lasso and Ridge regressions and the $\lambda$ and $\alpha$ parameters for Elastic Net are determined using the cross-validation strategy. For tree-based models (Random Forest and GBM) and Neural Network, cross-validation is mainly used to identify the best hyperparameters. We employ random grid search for hyperparameters selection, where parameters are uniformly sampled from the grid defined by the decision maker. Furthermore, we enable parallelism to set up models in parallel, aiming to accelerate the solution process.

We standardize the input vector to ensure that the mean and standard deviation were 0 and 1, respectively, except for tree-based models where normalization is not common practice. We are unable to test the Support Vector Machine (SVM) and XgBoost algorithms since H2O library does not provide regression-based SVM \citep{SupportV60} and XgBoost is not available on the Windows operating system \citep{XGBoost}.

In Section \ref{Implementation Details}, we discuss how our models are currently being used at GE Gas Power during its short-term and long-term operations. Then, in Section \ref{Statistical Analysis}, we present our statistical analysis for all seven models together with our findings.

\subsection{Implementation Details at GE Gas Power} \label{Implementation Details}

GE Gas Power is currently incorporating predictive modeling techniques into its daily business operations through a hybrid approach that combines the use of short, medium, and long-range availability dates. Before diving into the details of this approach, it's important to define each term:

\begin{enumerate}
    \item Short range availability,  also known as the "pick-up date, is the date on which the supplier notifies GE Gas Power, through the transportation management system (TMS), that the products are ready for pick-up. This date is considered the most accurate product availability date since it comes directly from the supplier, who provides it upon completing manufacturing, packaging, and staging for pick-up. Typically, this date will be available within 24-48 hours of pick-up.
    \item Medium-range availability, also known as the "supplier promised availability date," is the date on which the supplier promises to have the products ready to ship when GE Gas Power places a product order (PO). This date is an estimate based on the supplier's experience, workload, manufacturing cycle times, and shop capacity. It is subject to change as production starts and a more accurate date of production completion becomes known. This date will be available upon PO release, but it could be updated up to 30 days before the planned ready-to-ship date.
    \item Long-range availability, also known as the "forecasted product availability date," is the predicted date on which the supplier will be ready to ship, using historical data feeds coupled with a machine learning algorithm that forms the predictions. This date is predicted and made available once a PO is placed and the features of the PO are available to make the prediction. The prediction for this date will be forecasting availability dates out anywhere from 3-12 months into the future.
\end{enumerate}

 \begin{figure}[!htbp]
\centering
     \caption{Visualization of the logistics forecasted load profile}
     \includegraphics[page=1,width= 0.75 \linewidth,keepaspectratio=true]{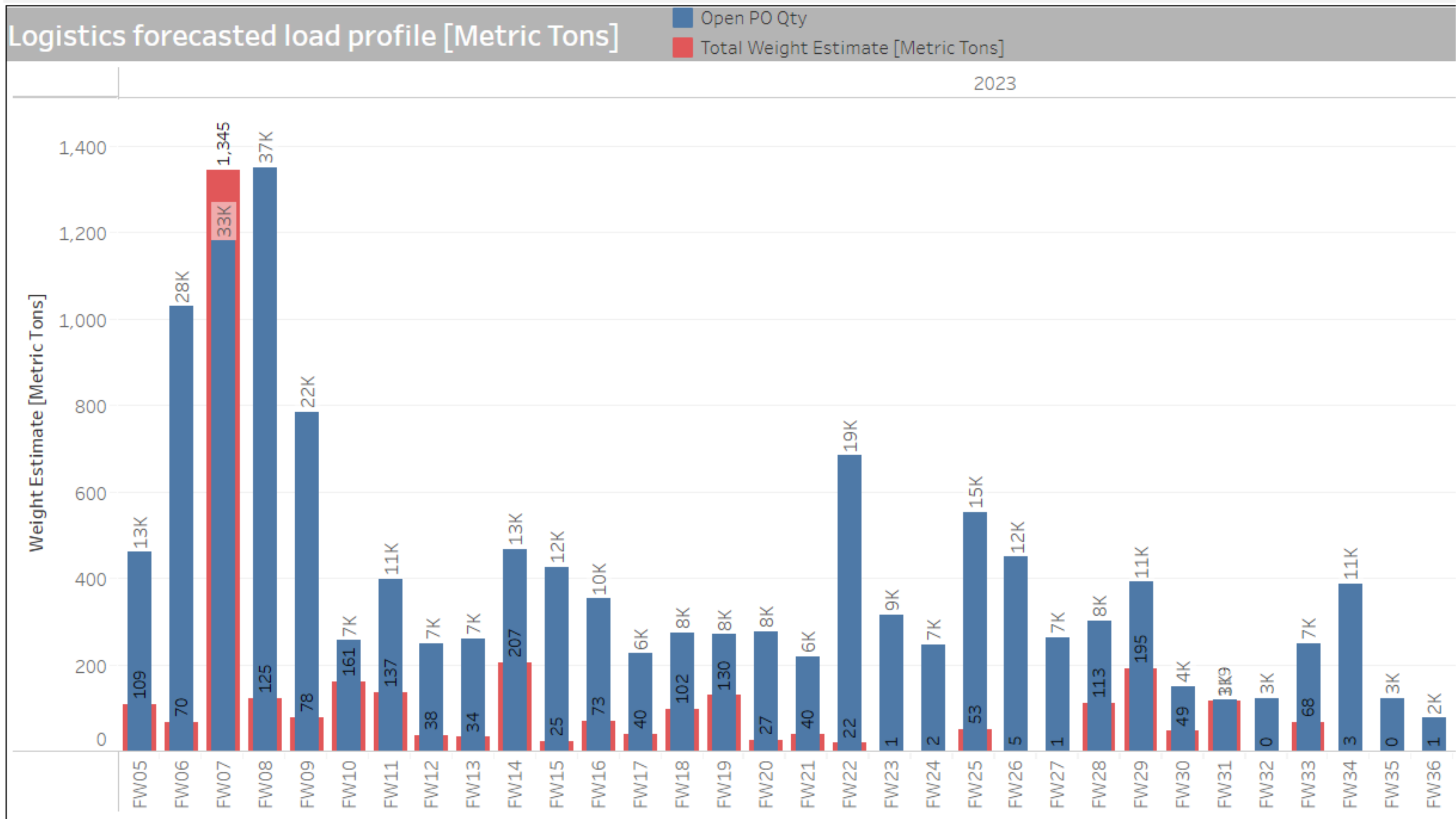}   \label{Fig3}
\end{figure}

The GE Gas Power team plans on leveraging all three date ranges as follows:

\begin{enumerate}
    \item Tactically:
        \begin{itemize}
            \item If the pick-up date is available, then the team will use it to plan the necessary assets for today, tomorrow, or next week that will be needed to move the products.
            \item If the supplier promised availability date is available but the pick-up date is not yet available, the team will plan based on the supplier's promised availability date. This plan is typically for 3-6 months in advance and is subject to change upon receiving the pick-up date notification from the supplier.
            \item Finally, if neither pick-up date nor supplier promised availability date are available, then the team will use the forecasted product availability date for long-term planning. However, similar to the supplier promised availability date, this date will be subject to change once the pick-up date notification is provided by the supplier.
        \end{itemize}
    \item Strategically:
        \begin{itemize}
            \item 	Both the forecasted product availability date and promised availability dates will be used in conjunction to estimate the future load profile over a given period (e.g., 12 months) for a specific shipping lane (e.g., city to city). As an example, Fig.~\ref{Fig3} shows a screenshot of a load profile for a particular lane within GE Gas Power.
            \item 	Based on the load profile analysis, key shipping lanes with high volumes and critical components needed to build GE Gas Power's products will be level loaded and have assets booked twice a week, weekly, or bi-weekly on a cadence that allows for standard work and avoids trying to time the booking based on estimated product availability dates. This approach will increase the reliability and resiliency of the supply chain.
            \item 	To ensure stable shipping costs, set shipping rates will be established for the key shipping lanes that are being level loaded. This approach will help lock down prices and hedge against price movements.
        \end{itemize}
\end{enumerate}

It is worth noting that the above sequence would take place at a purchase order line level and will be dynamic and continuously managed daily as orders move through the supply chain. This means that the team will be monitoring and adjusting the planning and scheduling processes as needed to ensure that the products are delivered on time and within budget.

\subsection{Statistical Analysis} \label{Statistical Analysis}
In this section, we  analyze the prediction results obtained from the regression models we tested. Our dataset consists of shipment information from the middle of 2019 to the end of 2022 and includes product orders only with destinations within the US. After extensive data cleaning, our dataset was reduced to 27,729 rows. To ensure accurate results, we split our dataset into training and testing data with a ratio of $80\%$ and $20\%$, respectively. We do not create a validation data due to the fact that we perform 5-fold cross validation as discussed earlier.

\begin{figure}[ht]
    \begin{minipage}{0.49\textwidth}
   	\includegraphics[scale=.5]{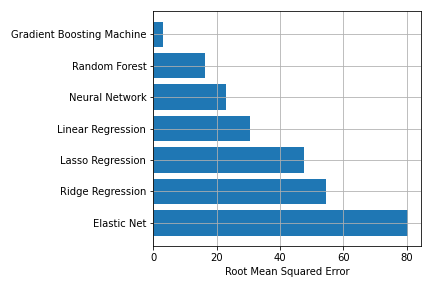}  
    \end{minipage}%
      \hfill
    \begin{minipage}{0.49\textwidth}
  	\includegraphics[scale=.5]{RMSE_train.png}  

    \end{minipage}
   
    \caption{RMSE for the models : (a) Training Data (left)  (b) Testing Data (right)}
     \label{RMSE_plot}
\end{figure}

\begin{figure}[!htbp]
    \begin{minipage}{0.49\textwidth}
   	\includegraphics[scale=.5]{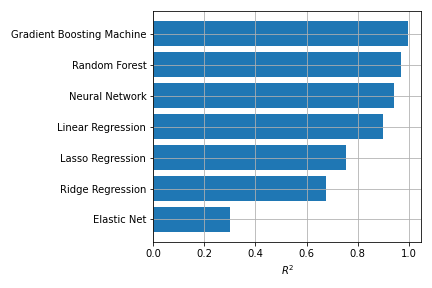}  
    \end{minipage}%
      \hfill
    \begin{minipage}{0.49\textwidth}
  	\includegraphics[scale=.5]{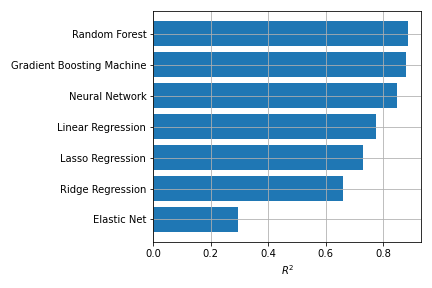} 

    \end{minipage}
   
  	\caption{R2 for the models : (a) Training Data (left)  (b) Testing Data (right)}
	\label{R2_plot}
\end{figure}

\begin{figure}[!htbp]

	\centering
	\begin{subfigure}[b]{0.4\linewidth}    
		\includegraphics[width=\linewidth]{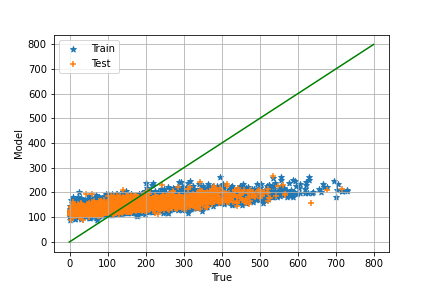}
		\caption{}
	\end{subfigure}
	\begin{subfigure}[b]{0.4\linewidth}    
		\includegraphics[width=\linewidth]{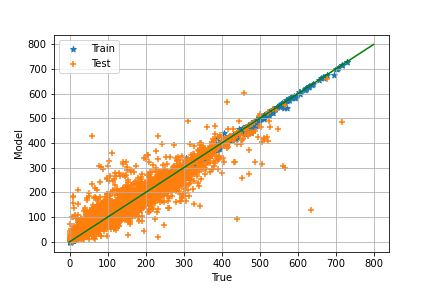}
		\caption{}
	\end{subfigure}
 
    \centering 
	\begin{subfigure}[b]{0.4\linewidth}    
		\includegraphics[width=\linewidth]{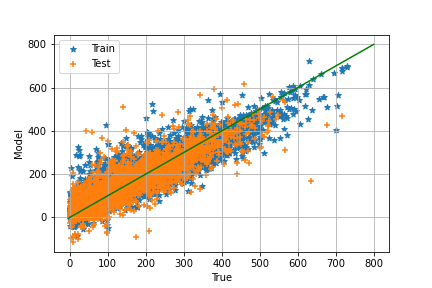}
		\caption{}
	\end{subfigure}
	\begin{subfigure}[b]{0.4\linewidth}    
		\includegraphics[width=\linewidth]{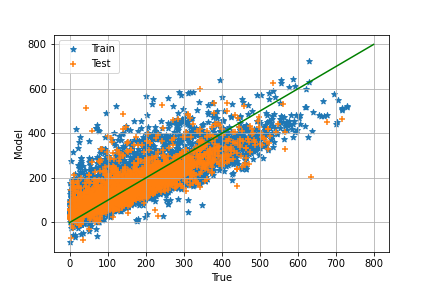}
		\caption{}
	\end{subfigure}

    \centering 
	\begin{subfigure}[b]{0.4\linewidth}    
		\includegraphics[width=\linewidth]{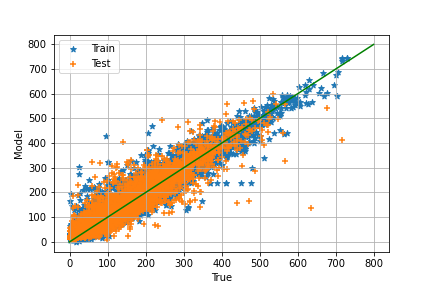}
		\caption{}
	\end{subfigure}
	\begin{subfigure}[b]{0.4\linewidth}
		\includegraphics[width=\linewidth]{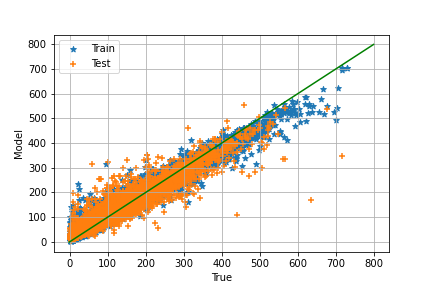}
		\caption{}
	\end{subfigure}

    \centering
    \begin{subfigure}[b]{0.4\linewidth}
		\includegraphics[width=\linewidth]{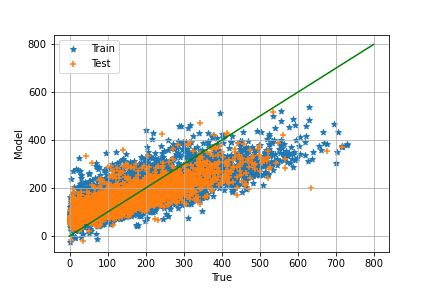}
		\caption{}
	\end{subfigure}
 
	\caption{45 Degree Plot for the models : (a) Elastic net  (b) Gradient Boosting Machine (c) Linear Regression (d) Lasso Regression (e) Neural Net (f) Random Forest (g) Ridge Regression}
	\label{45deg_plot}
\end{figure}

\begin{figure}[!htbp]

	\centering
	\begin{subfigure}[b]{0.4\linewidth}    
		\includegraphics[width=\linewidth]{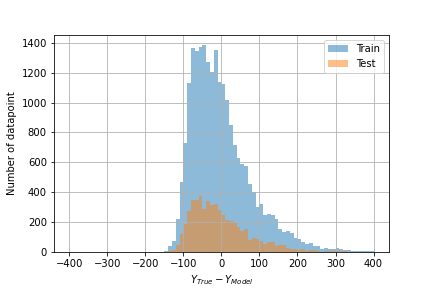}
		\caption{}
	\end{subfigure}
	\begin{subfigure}[b]{0.4\linewidth}    
		\includegraphics[width=\linewidth]{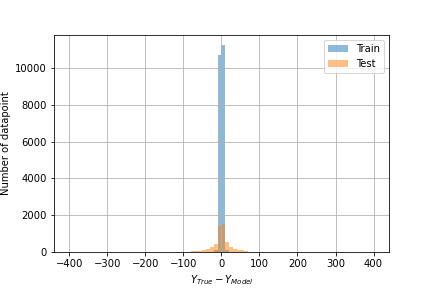}
		\caption{}
	\end{subfigure}
 
    \centering 
	\begin{subfigure}[b]{0.4\linewidth}    
		\includegraphics[width=\linewidth]{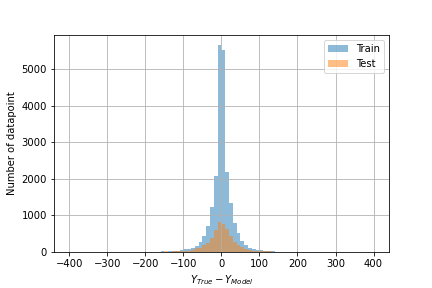}
		\caption{}
	\end{subfigure}
	\begin{subfigure}[b]{0.4\linewidth}    
		\includegraphics[width=\linewidth]{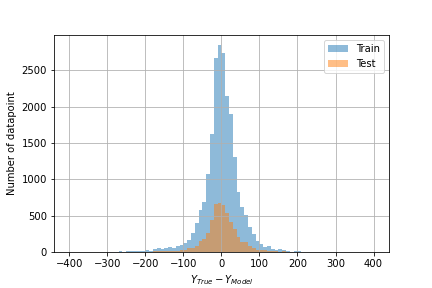}
		\caption{}
	\end{subfigure}

    \centering 
	\begin{subfigure}[b]{0.4\linewidth}    
		\includegraphics[width=\linewidth]{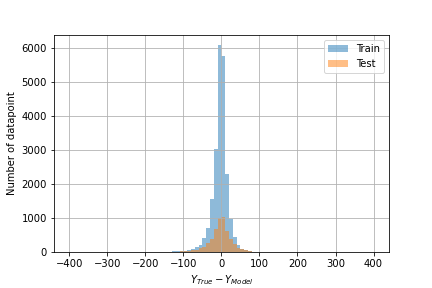}
		\caption{}
	\end{subfigure}
	\begin{subfigure}[b]{0.4\linewidth}
		\includegraphics[width=\linewidth]{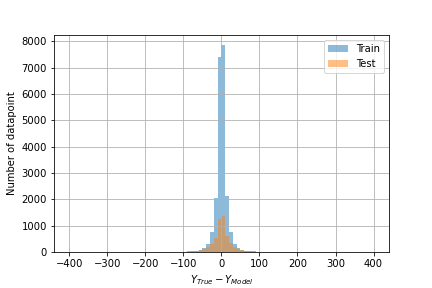}
		\caption{}
	\end{subfigure}

    \centering
    \begin{subfigure}[b]{0.4\linewidth}
		\includegraphics[width=\linewidth]{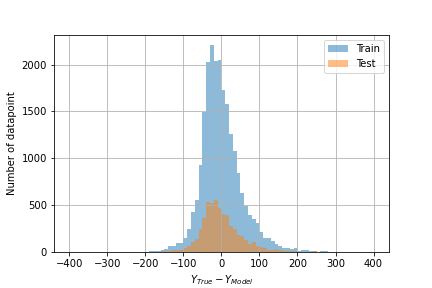}
		\caption{}
	\end{subfigure}
 
	\caption{Histogram Plot for the models : (a) Elastic net  (b) Gradient Boosting Machine (c) Linear Regression (d) Lasso Regression (e) Neural Net (f) Random Forest (g) Ridge Regression}
	\label{Hist_plot}
\end{figure}

The results from the models are analyzed through four key plots: a) the root mean squared error (RMSE) comparison plot (Figure \ref{RMSE_plot}), b) $R^2$ comparison plot (Figure  \ref{R2_plot}), c) 45 degree plot (Figure  \ref{45deg_plot}) , and d) prediction histogram (Figure \ref{Hist_plot}). The RMSE and $R^2$  comparison plots presents the RMSE and   $R^2$   for the training and testing data across the seven models considered for the study here, respectively.  Additionally, the performance of the models are also analyzed through the 45 degree plots of the training and testing data. Taking it a step further, the prediction difference is  calculated  as a histogram for each of the model to further understand the trends in performance. The prediction difference is calculated as $y_{true} - y_{pred}$. 


Based of the four plots given, we make the following observations and inferences:
\begin{itemize}
    \item From the testing RMSE and $R^2$ values of the seven models, it can be identified the best performing model is Random Forest (RF) (see Fig.~\ref{RMSE_plot} and Fig.~\ref{R2_plot}). However, Gradient Boosting Machine (GBM) and Neural Networks (NN) are not far behind showing only a difference of approximately one and four days, respectively in their RMSE values. The rest of the regression models  start showing higher differences when compared to the top three models. This is expected because out of the seven models, GBM, RF, and NNs have the highest non-linearity embedding in them. However, Linear Regression, Lasso Regression, Ridge Regression and Elastic Net all represent a class first order regression models with different regularization terms. 
    \item Even among the four first regression models, the Linear Regression model stands out as the best model. This is due to the lack of regularization term that enables less regularized model capable of modeling the non-linear phenomenon under study. On the other hand, Eastic Net which has two regularization terms is the worst performing model due to its inability to model complex response surfaces. This is example where regularization might not always serve the benefit of a the data-driven model. In the steps to prevent over-fitting and ensure generalization, can lead to underfit models.
    \item The conclusions drawn from the comparison of RMSE is also corroborated by the compiled $R^2$ the conclusions drawn on regularization drawn from the RMSE and $R^2$ values of the four models.
    \item The superiority of RF, GBM, and NNs in performance can be observed through the 45-degree plots, as depicted in Fig.~\ref{45deg_plot}. The plots show a better overlap with the 45-degree line and a narrower spread, indicating a higher degree of accuracy.
    \item The histogram plot of the prediction difference provides further insight into the nuances of the models, as shown in Fig.~\ref{Hist_plot}. With the exception of Elastic Net, Ridge Regression, and Lasso Regression, all the histograms exhibit a symmetric behavior. Furthermore, the histograms are more or less centered around zero, which is in line with expectations. The regularized first-order linear models, on the other hand, display higher spreads. The GBM model shows the minimum spread of the histogram and the highest concentration of data points around zero, followed by RF and NN.
\end{itemize}

The thorough analysis of seven models conducted for the current study led to some noteworthy observations. Firstly, the tree-based ensemble methods, including Random Forest (RF) and Gradient Boosting Machines (GBM), exhibited superior performance compared to the other models assessed. The accuracy of these models was evident from their narrow spread and better overlap with the 45-degree line.

On the other hand, Elastic Net showed the poorest performance among the models assessed, as revealed by the histogram plot of the prediction difference. The histogram plot revealed a skewed distribution, indicating that the model had difficulty predicting accurately for certain data points. Another interesting finding was that the models were hindered by regularization, which negatively impacted their performance. The first-order linear models, such as Ridge Regression and Lasso Regression, exhibited higher spreads in their histograms compared to other models, further supporting the adverse effect of regularization. Moreover, the analysis highlighted the importance of considering the non-linearity of the data. The regularized models struggled to capture the non-linear patterns in the data, leading to decreased accuracy. This observation aligns with the notion that tree-based ensemble methods are better suited for capturing non-linear relationships in the data.

\begin{figure}[!htbp]
 \centering	\includegraphics[scale=0.5]{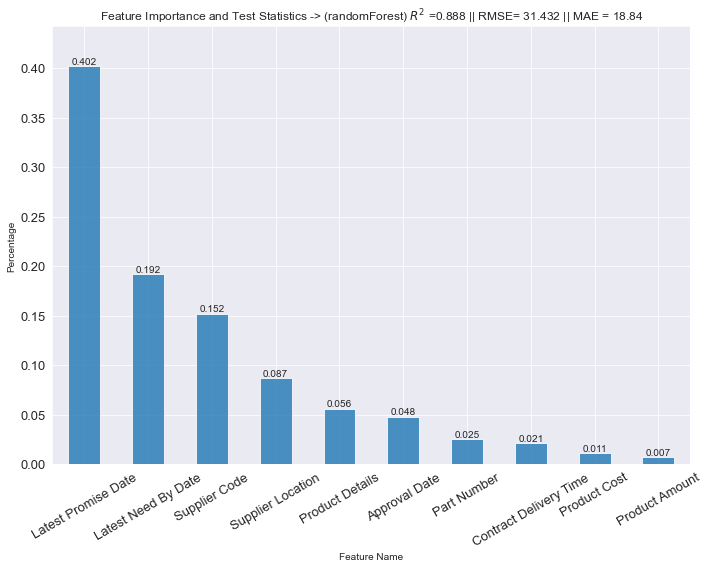}
	\caption{Feature importance plot - Random Forest }
	\label{rf_feature}
\end{figure}

\begin{figure}[!htbp]
 
 \centering
\includegraphics[scale=0.5]{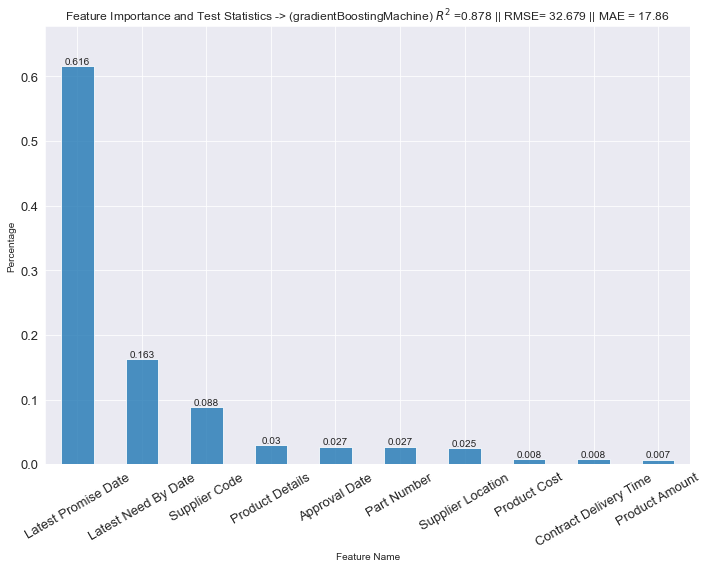}
	\caption{Feature importance plot - Gradient Boosting Machine}
	\label{gbm_feature}
\end{figure}

As detailed in Section \ref{Methodology}, tree ensemble models are useful learning tools for measuring feature importance. This is calculated based on the average decrease in node impurity, which is weighted by the probability of reaching the corresponding node (i.e., feature) in the forest. The percentage values of feature importance for both RF and GBM models are presented in  Fig.~\ref{rf_feature} and Fig.~\ref{gbm_feature}, respectively.  We also present the values of $R^2$, RMSE, and mean-absolute value (MAE) for the test data in our figures. 

Both RF and GBM models reveal that "Latest Promise Date", "Latest Need By Date", and "Supplier Code" are the most significant features in the decision-making process for predicting product availability dates. This finding is intuitive since these dates are determined by GE Power and suppliers when creating a product order. Interestingly, the "Product Amount" and "Product Cost" features do not seem to have a significant impact on the availability dates. While we initially experimented with removing these features from the models, we found that doing so led to decreased test statistic values, indicating that they do in fact play a crucial role in the predictions due to the non-linear relationships within the data. This also explains why simpler regression models perform poorly in capturing the true values of product availability dates.

In contrast, the "Contract Delivery Date" feature has an importance of less than $0.1\%$, which may seem counterintuitive since it is expected to have a positive correlation with product availability dates. However, we observed a high percentage of missing data in this feature, accounting for roughly $50\%$ of the total dataset. This can be attributed to the difficulty in obtaining delivery date information from suppliers and the lack of systematic preservation of historical data. As the data pipeline becomes more accurate and consistent, the importance of contract delivery dates may increase.

Suppliers often manufacture and ship products with different characteristics, and despite the company signing independent contracts per product order with agreed-upon dates (such as latest promised date or latest need by date), the suppliers may only notify the business of their readiness to ship all orders at once. As a result, the business is compelled to pick up these orders on the same day, regardless of the original promised date. Differently, the business might decide to pick up orders on the same day even if the products are prepared at different times by a supplier which is still stored as the same availability date in the system. This non-linear behavior poses a significant challenge in predicting the product availability dates accurately, and it may be a primary reason why the machine learning algorithms tested in this study are not producing even better generalization results. Therefore, it is important to acknowledge the limitations of the current algorithms and continue to explore other approaches to improve prediction accuracy in a future study.
\section{Conclusion} \label{Conclusion}

In this study, we employed a range of regression models, including Simple Regression, Lasso Regression, Ridge Regression, Elastic Net, Random Forest (RF), Gradient Boosting Machine (GBM), and Neural Network, to predict the availability date of product orders at supplier locations for GE Gas Power. Our computational experiments demonstrated that tree-based learning algorithms, such as RF and GBM, yielded the best performance, outperforming other models in terms of test error. The accuracy of these predictions is crucial for inbound logistics planning, and the results will be integrated into a mathematical optimization model to generate annual shipment plans from Asia to North America.

In future research, it would be interesting to investigate the impact of seasonality and time on availability date predictions. As supplier performance can improve or deteriorate over time, incorporating time series analysis into our predictions may be valuable, especially, as the size of the historical data expands in the following years. Furthermore, researchers might explore the potential of using advanced deep learning techniques such as Long Short-Term Memory  and transformers in a future work. These approaches have shown promising results in time series forecasting and may improve the accuracy of our predictions.


\end{document}